\documentclass[conference,letterpaper]{IEEEtran}
\IEEEoverridecommandlockouts

\usepackage{amsmath,amssymb}
\usepackage{graphicx}
\usepackage{textcomp}
\usepackage[table]{xcolor}
\usepackage[hidelinks]{hyperref}
\begin{document}

\title{
Toolkit for Facial Deidentification\\[2pt]
\small\itshape (Toolkit para Desidentificaci\'on Facial)
\thanks{This work was presented as a poster at the 6th IEEE International
Conference on BioInspired Processing (BIP 2024), poster session, Liberia,
Costa Rica, December 4--6, 2024.}}

\newcommand{\AName}[1]{\hbox to 0.5\hsize{\hfill\makebox[83mm][c]{#1}\hfill}}\newcommand{\AAffil}[1]{\hbox to 0.5\hsize{\hfill\parbox[t]{83mm}{\centering#1}\hfill}}
\author{
\IEEEauthorblockN{\AName{Esteban Leiva-Montenegro\IEEEauthorrefmark{1}}}
\IEEEauthorblockA{\AAffil{\IEEEauthorrefmark{1}Instituto Tecnol\'ogico de
Costa Rica, Costa Rica\\
\textit{estebanlm2803@gmail.com}\\
\footnotesize ORCID: \url{https://orcid.org/0009-0004-3523-0057}}}

\and

\IEEEauthorblockN{\AName{Vitomir \v{S}truc\IEEEauthorrefmark{3}}}
\IEEEauthorblockA{\AAffil{\IEEEauthorrefmark{3}University of Ljubljana,
Ljubljana, Slovenia\\
\textit{vitomir.struc@fe.uni-lj.si}\\
\footnotesize ORCID: \url{https://orcid.org/0000-0002-3385-5780}}}

\and

\IEEEauthorblockN{\AName{Manfred Gonzalez-Hernandez\IEEEauthorrefmark{3}}}
\IEEEauthorblockA{\AAffil{\IEEEauthorrefmark{3}University of Ljubljana,
Ljubljana, Slovenia\\
\textit{gonzalezhernandez.manfred@gmail.com}\\
\footnotesize ORCID: \url{https://orcid.org/0000-0002-5408-7901}}}

\and

\IEEEauthorblockN{\AName{Matthieu Pillot\IEEEauthorrefmark{2}}}
\IEEEauthorblockA{\AAffil{\IEEEauthorrefmark{2}CERI/LARIS, ENSAM, Universit\'e
Paris-Est Cr\'eteil, France\\
\textit{matthieu.pillot@ensea.fr}\\
\footnotesize ORCID: \url{https://orcid.org/0009-0006-3179-9152}}}

\and

\IEEEauthorblockN{\AName{Peter Peer\IEEEauthorrefmark{3}}}
\IEEEauthorblockA{\AAffil{\IEEEauthorrefmark{3}University of Ljubljana,
Ljubljana, Slovenia\\
\textit{peter.peer@fri.uni-lj.si}\\
\footnotesize ORCID: \url{https://orcid.org/0000-0001-9744-4035}}}

\and

\IEEEauthorblockN{\AName{Bla\v{z} Meden\IEEEauthorrefmark{3}}}
\IEEEauthorblockA{\AAffil{\IEEEauthorrefmark{3}University of Ljubljana,
Ljubljana, Slovenia\\
\textit{blaz.meden@fri.uni-lj.si}\\
\footnotesize ORCID: \url{https://orcid.org/0000-0002-1690-4798}}}}

\maketitle

\begin{abstract}
This work presents a novel open-source toolkit designed for the evaluation of
face deidentification techniques with available evaluation models. The
framework provides a complete evaluation pipeline, including preprocessing,
deidentification, and performance assessment, based on data utility, image
quality, and identity protection metrics. It supports a variety of datasets,
enabling researchers to easily benchmark new deidentification techniques.
Results show the effectiveness of some techniques on maintaining privacy while
preserving image utility. The toolkit is aimed at unifying future research by
providing a standard evaluation framework for privacy-enhancing technologies in
facial biometrics. Implementation is available at
\url{github.com/blazm/deid-toolkit}.
\end{abstract}


\section{Introduction}
Biometric data, such as facial images, is inherently tied to individuals and
can reveal a substantial amount of personal information~\cite{meden2021survey}.
Today, there is a wide range of automated methods capable of extracting features
from facial images, regardless of variations and noise in the data. This feature
extraction can potentially expose sensitive information about individuals and
enable face recognition within large databases, thereby compromising a
fundamental human right---privacy~\cite{meden2023cpp}, as protected under the
Human Rights Act of 1998, Article 8: Right
to Privacy~\cite{humanrightsact}.

\begin{figure}[htb]
  \centering
  \includegraphics[width=\columnwidth]{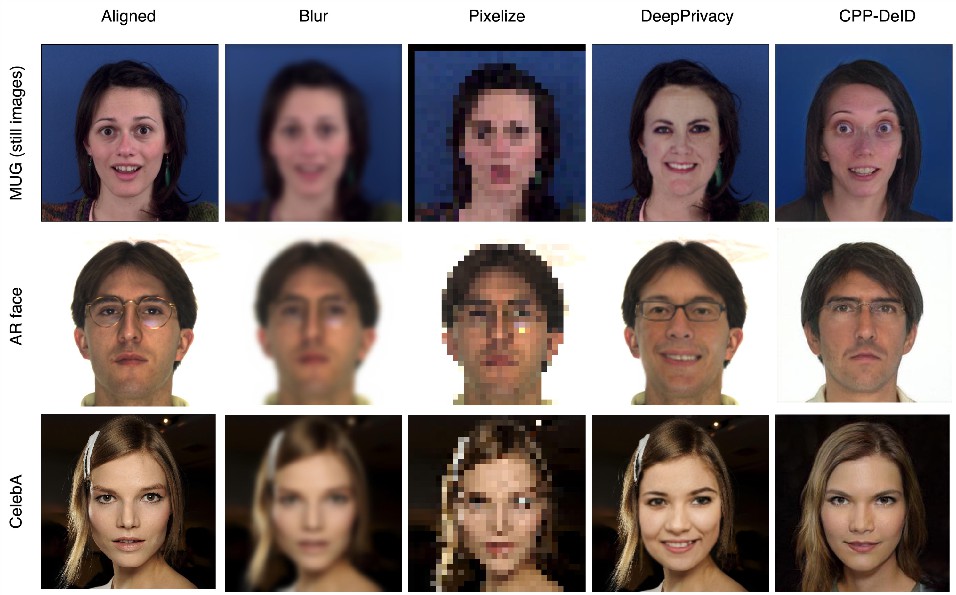}
  \caption{Examples of visual results before and after deidentification---with
  reference images and the results visualized along with the rest of the
  deidentification techniques available in the toolkit.}
  \label{fig:examples}
\end{figure}

Face deidentification aims to safeguard privacy by concealing or removing
identity-related information, making it difficult for both human observers and
machine learning models to infer the subject's identity~\cite{meden2023cpp}. In
addition to privacy protection, another critical goal is utility preservation.
While identity-specific features are obfuscated, the rest of the facial features
must remain intact to retain the image's utility. The significance of facial
deidentification has gained considerable attention in recent years, driven by
the need to preserve privacy in facial images across various applications.
Numerous methods have been proposed, leveraging a range of machine learning
techniques to achieve privacy protection~\cite{li2021latent, hu2022cvpr,
yang2021iccv, uchida2022dedim, meden2023cpp}. Some of these techniques are
evaluated using various models (classifiers, recognition models, metrics) to
ensure that the utility of the deidentified data is maintained and that privacy
is protected by preventing recognition of individuals in the processed images
(see \autoref{fig:examples} for visual examples of deidentification methods within
the toolkit). For more detailed descriptions and explanations of existing
deidentification techniques and evaluation methodology, we refer the reader to
the related literature~\cite{meden2021survey, meden2023cpp}.

Beyond privacy protection, deidentification techniques must ensure high realism
to remain artifact-free, especially for human-centered applications like
attribute recognition, where image realism is crucial~\cite{meden2023cpp}.
Additionally, state-of-the-art (SOTA) models need to generate diverse,
deidentified outputs, particularly when handling multiple
subjects~\cite{nousi2020ae}. Proper data preparation and configuration for model
inference, as well as evaluating data utility and identity verification, are
critical for advancing research in facial deidentification. The main
contribution of this work is a centralized solution with data preparation
techniques and SOTA deidentification models, supported by comprehensive
performance metrics and protocols.

\section{Methodology}
In this section, we outline the proposed toolkit's structure, covering the data
preparation techniques (face detection and face alignment via detected
landmarks) and both SOTA and classical methods for facial deidentification.
Techniques range from simple blurring to advanced deep learning-based facial
modifications. We also present the evaluation and verification metrics
integrated to assess the performance of these techniques. This toolkit is the
first of its kind for open evaluation of facial deidentification techniques.
The toolkit features a user-friendly command-line interface, enabling users to
browse available datasets, techniques, and metrics, and to perform specific
operations such as dataset preprocessing, applying deidentification techniques,
or running evaluation metrics on deidentified results.

\begin{figure}[htb]
  \centering
  \includegraphics[width=\columnwidth]{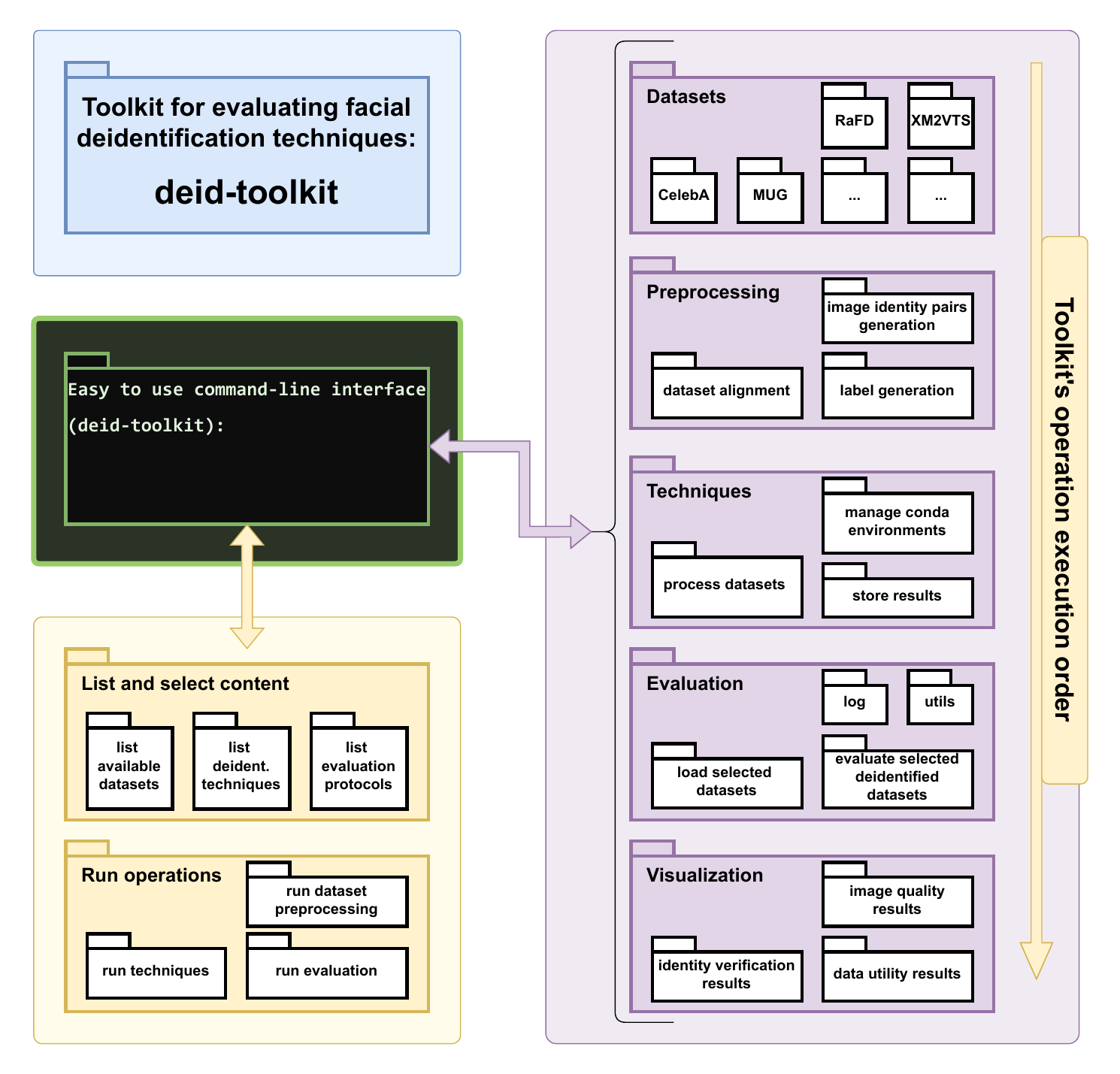}
  \caption{The architecture design of the proposed facial deidentification
  toolkit.}
  \label{fig:architecture}
\end{figure}

The toolkit is organized into five main categories: \emph{datasets},
\emph{preprocessing}, \emph{techniques}, \emph{evaluation}, and
\emph{visualization}, as shown in \autoref{fig:architecture}. Users can select
datasets, and preprocessing includes image preparation and metadata generation.
The techniques package manages the deidentification algorithms, handling
dependencies and encapsulating environments, while also processing datasets and
storing results using consistent naming conventions for easier evaluation.
Once deidentification is complete, users can execute the evaluation package,
which offers protocols for re-identification attempts, facial attribute
estimation, data utility, image quality, and diversity assessments. Users
select the desired metrics, and upon completion, results can be visualized
through plots, metric visualizations, and grid-based displays of deidentified
images to evaluate qualitative outcomes. The implementation is available at
\url{github.com/blazm/deid-toolkit}.

\section{Results}
We provide open protocols to evaluate various face deidentification algorithms.
While datasets are not included in the toolkit due to licensing restrictions,
they can be adapted for use as long as they have annotated attribute labels.
For this study, we followed the established literature to select suitable facial
image datasets~\cite{meden2021survey}, with MUG~\cite{aifanti2008mug},
Extended Cohn-Kanade (CK+)~\cite{cohn2000ck}, and ArFace~\cite{martinez1998ar}
being key examples.

The MUG dataset contains sequences from 86 subjects displaying seven emotions
and includes gender annotations. We processed MUG into a ``MUG-still'' variant,
where only still frames showing peak expressions were retained for analysis
(988 frames). CK+ and ArFace datasets include 5876 and 3128 annotated facial
images, respectively. CK+ provides expression labels, while ArFace provides
gender labels. All the datasets were preprocessed using the toolkit's internal
preprocessing protocols, including detection and alignment. The reported
results cover four categories: identity verification after
deidentification via AdaFace~\cite{kim2022adaface} (\autoref{tab:quality}, right),
gender and expression preservation evaluation via the DAN~\cite{tautkute2019dan},
HSEmotion~\cite{hsemotion}, DeepFace~\cite{taigman2014deepface}, and
ResNet-18~\cite{abhishek2022resnet18} models (\autoref{tab:utility}), image
quality evaluation metrics (\autoref{tab:quality}, left), and the qualitative
evaluation with visual examples (\autoref{fig:examples}).

\begin{table*}[htb]
  \centering
  \caption{Accuracy of deidentification techniques in preserving data utility,
  measured by facial expression classification (7 classes, using
  DAN~\cite{tautkute2019dan} and HSEmotion~\cite{hsemotion}) and gender
  prediction (using DeepFace~\cite{taigman2014deepface} and
  ResNet-18~\cite{abhishek2022resnet18}). DAN's performance was excluded for
  CPP-DeID, as it served as the guidance model. Evaluation was conducted on the
  MUG-still, CK+, and ArFace datasets. Deidentification techniques are listed
  in the first column and the evaluation classification models are listed in
  the second row (in gray).}
  \label{tab:utility}
  \footnotesize
  \renewcommand{\arraystretch}{1.0}
    \resizebox{0.95\textwidth}{!}{
  \begin{tabular}{l | cccc | cc | cc}
    \hline
    & \multicolumn{4}{c}{\textbf{MUG (still) dataset}}
    & \multicolumn{2}{c}{\textbf{CK+ dataset}}
    & \multicolumn{2}{c}{\textbf{ArFace dataset}} \\
     \textbf{Techniques}
    & \cellcolor{gray!25}DAN (expr.) {\scriptsize$\uparrow$}
    & \cellcolor{gray!25}HSEmotion (expr.) {\scriptsize$\uparrow$}
    & \cellcolor{gray!25}DeepFace (gd) {\scriptsize$\uparrow$}
    & \cellcolor{gray!25}ResNet-18 (gd) {\scriptsize$\uparrow$}
    & \cellcolor{gray!25}DAN (expr.) {\scriptsize$\uparrow$}
    & \cellcolor{gray!25}HSEmotion (expr.) {\scriptsize$\uparrow$}
    & \cellcolor{gray!25}DeepFace (gd) {\scriptsize$\uparrow$}
    & \cellcolor{gray!25}ResNet-18 (gd) {\scriptsize$\uparrow$} \\
    \hline
    AMT-GAN      & $76.98\%$ & $80.73\%$ & $90.57\%$ & $91.68\%$
                  & $78.19\%$ & $81.88\%$
                  & $88.44\%$ & $91.52\%$ \\
    Croft et al. & $13.18\%$ & $12.58\%$ & $59.63\%$ & $41.08\%$
                  & $18.64\%$ & $17.09\%$
                  & $78.04\%$ & $35.42\%$ \\
    Blur         & $60.34\%$ & $56.59\%$ & $81.85\%$ & $54.97\%$
                  & $69.40\%$ & $74.87\%$
                  & $85.65\%$ & $74.58\%$ \\
    CleanIR      & $43.91\%$ & $42.80\%$ & $71.40\%$ & $84.08\%$
                  & $56.42\%$ & $56.78\%$
                  & $78.71\%$ & $81.23\%$ \\
    CPP-DeID     & / & $66.63\%$ & $86.71\%$ & $94.42\%$
                  & / & $67.39\%$
                  & $86.35\%$ & $88.84\%$ \\
    KSameNet     & $71.50\%$ & $67.34\%$ & $58.22\%$ & $56.09\%$
                  & $49.17\%$ & $51.48\%$
                  & $72.03\%$ & $55.08\%$ \\
    Pixelize     & $39.76\%$ & $44.42\%$ & $62.17\%$ & $59.33\%$
                  & $52.31\%$ & $48.12\%$
                  & $83.38\%$ & $65.22\%$ \\
    \hline
  \end{tabular}}
\end{table*}

\begin{table*}[htb]
  \centering
  \small
  \caption{Baseline performances in terms of computing image quality metrics,
  obtained on deidentified and source image pairs---including MSE,
  LPIPS~\cite{zhang2018lpips}, SSIM~\cite{wang2004ssim}, and
  MS-SSIM~\cite{wang2004ssim} (left). Verification experiment results in terms
  of the reported Area Under the ROC Curve (AUC~\cite{bradley1997roc}) (right).
  The evaluation was performed on the MUG-still, CK+, and ArFace datasets.}
  \label{tab:quality}
  \footnotesize
  \renewcommand{\arraystretch}{1.0}
  \resizebox{0.95\textwidth}{!}{
  \begin{tabular}{l | cccc | ccc}
    \hline
    & \multicolumn{4}{c}{\textbf{image quality}} & \multicolumn{3}{c}{\textbf{verification (AUC {\scriptsize$\downarrow$})}} \\
     \textbf{Techniques}
    & \cellcolor{gray!25}MSE $\downarrow$
    & \cellcolor{gray!25}LPIPS $\downarrow$
    & \cellcolor{gray!25}SSIM $\uparrow$
    & \cellcolor{gray!25}MS-SSIM $\uparrow$
    & \cellcolor{gray!25}MUG (still)
    & \cellcolor{gray!25}CK+
    & \cellcolor{gray!25}ArFace \\
    \hline
    AMT-GAN      & $0.20 \pm 0.06$ & $0.43 \pm 0.09$ & $0.51 \pm 0.12$ & $0.59 \pm 0.11$
                  & $0.99 \pm 0.00$ & $0.99 \pm 0.00$ & $0.74 \pm 0.01$ \\
    Croft et al. & $1.04 \pm 0.14$ & $0.67 \pm 0.02$ & $0.35 \pm 0.04$ & $0.24 \pm 0.13$
                  & $0.52 \pm 0.01$ & $0.50 \pm 0.00$ & $0.51 \pm 0.00$ \\
    Blur         & $0.01 \pm 0.00$ & $0.57 \pm 0.04$ & $0.71 \pm 0.05$ & $0.82 \pm 0.03$
                  & $0.99 \pm 0.00$ & $0.99 \pm 0.00$ & $0.72 \pm 0.00$ \\
    CleanIR      & $0.01 \pm 0.00$ & $0.23 \pm 0.02$ & $0.81 \pm 0.04$ & $0.90 \pm 0.02$
                  & $0.77 \pm 0.01$ & $0.84 \pm 0.00$ & $0.64 \pm 0.00$ \\
    CPP-DeID     & $0.03 \pm 0.01$ & $0.19 \pm 0.03$ & $0.66 \pm 0.05$ & $0.73 \pm 0.06$
                  & $0.95 \pm 0.01$ & $0.99 \pm 0.00$ & $0.72 \pm 0.01$ \\
    KSameNet     & $1.15 \pm 0.15$ & $0.72 \pm 0.04$ & $0.38 \pm 0.05$ & $0.36 \pm 0.12$
                  & $0.50 \pm 0.01$ & $0.51 \pm 0.00$ & $0.52 \pm 0.01$ \\
    Pixelize     & $0.05 \pm 0.01$ & $0.74 \pm 0.06$ & $0.59 \pm 0.05$ & $0.60 \pm 0.04$
                  & $0.99 \pm 0.00$ & $0.99 \pm 0.00$ & $0.73 \pm 0.00$ \\
    \hline
  \end{tabular}}
\end{table*}

\section{Conclusion}
The purpose of the presented toolkit for facial deidentification is to promote
open evaluation and to enable the use of structured and automated evaluation
protocols to measure the performances of currently available face
deidentification techniques. Additionally, the broader utility of the proposed
toolkit is to offer new researchers in this field the option to easily and
objectively compare the performance of newly developed facial
privacy-protecting techniques with existing baselines. The toolkit will be
regularly updated with new metrics to maintain a robust pipeline for face
deidentification.

\section*{Resumen (Espa\~{n}ol)}
\label{sec:resumen}
Este trabajo presenta una novedosa herramienta de c\'odigo abierto dise\~{n}ada
para la evaluaci\'on de t\'ecnicas de desidentificaci\'on facial utilizando
modelos de evaluaci\'on disponibles. El marco de trabajo proporciona una
canalizaci\'on completa de evaluaci\'on, que incluye preprocesamiento,
desidentificaci\'on y evaluaci\'on de rendimiento basada en la utilidad de los
datos, la calidad de la imagen y las m\'etricas de protecci\'on de identidad.
Soporta una variedad de conjuntos de datos, permitiendo a los investigadores
evaluar f\'acilmente nuevas t\'ecnicas de desidentificaci\'on. Los resultados
muestran la efectividad de algunas t\'ecnicas para mantener la privacidad
mientras se preserva la utilidad de la imagen. La herramienta est\'a destinada
a unificar investigaciones futuras al proporcionar un marco de trabajo
est\'andar de evaluaci\'on para las tecnolog\'ias que mejoran la privacidad en
biometr\'ia facial.

\bibliographystyle{IEEEtran}
\bibliography{references}

\end{document}